# Partial Order MCMC for Structure Discovery in Bayesian Networks

**Teppo Niinimäki**, **Pekka Parviainen** and **Mikko Koivisto**
Helsinki Institute for Information Technology HIIT, Department of Computer Science
University of Helsinki, Finland
{teppo.niinimaki, pekka.parviainen, mikko.koivisto}@cs.helsinki.fi

## Abstract

We present a new Markov chain Monte Carlo method for estimating posterior probabilities of structural features in Bayesian networks. The method draws samples from the posterior distribution of partial orders on the nodes; for each sampled partial order, the conditional probabilities of interest are computed exactly. We give both analytical and empirical results that suggest the superiority of the new method compared to previous methods, which sample either directed acyclic graphs or linear orders on the nodes.

## 1 INTRODUCTION

Learning the structure of a Bayesian network—that is, a directed acyclic graph (DAG)—from given data is an extensively studied problem. The problem is very challenging as it is intrinsically a model selection problem with a large degree of nonidentifiability. This, in particular, makes the Bayesian paradigm (Cooper and Herskovits, 1992; Heckerman et al., 1995; Madigan and York, 1995) an appealing alternative to classical maximum-likelihood or independence testing based approaches. In the Bayesian approach, the goal is to draw useful summaries of the posterior distribution over the DAGs. While sometimes a single maximum-a-posteriori DAG (MAP-DAG) provides an adequate summary, often the posterior probability of any single DAG or the associated Markov equivalence class is very small: more informative characterizations of the posterior distribution are obtained in terms of small subgraphs—such as individual arcs—that do have a relatively high posterior probability (Friedman and Koller, 2003).

Implementing the Bayesian paradigm seems, however, computationally hard. The fastest known algorithms for finding a MAP-DAG or computing the posterior probabilities of all potential arcs require time exponential in the number of nodes (Ott and Miyano, 2003; Koivisto and Sood, 2004; Tian and He, 2009), and are thus feasible only for around 30 nodes or fewer. For principled approximations, the popular Markov chain Monte Carlo (MCMC) method has been employed in various forms. Madigan and York (1995) implemented a relatively straightforward *Structure MCMC* that simulates a Markov chain by simple Metropolis–Hastings moves, with the posterior of DAGs as the target (stationary) distribution. Later, Friedman and Koller (2003) showed that mixing and convergence of the Markov chain can be considerably improved by *Order MCMC* that does not operate directly with DAGs but in the space of linear orders of the nodes. Such improvements are quite expected, since the space of linear orders is much smaller than the space of DAGs and, furthermore, the target (posterior) distribution over orders is expected to be smoother.

While Order MCMC is arguably superior to Structure MCMC, it raises two major questions concerning its limitations. First, Order MCMC assumes that the prior over DAGs is of a particular restricted form, termed *order-modular* (Friedman and Koller, 2003; Koivisto and Sood, 2004). Unfortunately, with an order-modular prior one cannot represent some desirable priors, like ones that are uniform over any Markov equivalence class of DAGs; instead, order-modularity forces DAGs that have a larger number of topological orderings to have a larger prior probability. Consequently, some modellers are not expected to find the order-modular prior and, hence, the order MCMC scheme, entirely satisfactory. This has motivated some researchers to enhance the structure MCMC scheme that allows arbitrary priors over DAGs. Eaton and Murphy (2007) use exact computations (which assume order-modularity) to obtain an efficient proposal distribution in an MCMC scheme in the space of DAGs. While this *hybrid MCMC* is very successful in correcting the "prior bias" of Order MCMC or of exact computations alone, the method does not scale to larger networks due to its exponential time complexity. Grzegorczyk and Husmeier (2008) throw away exponential-time exact computations altogether and, instead, introduce a new edge reversal move in the space of DAGs. Their experimental results suggest that, regarding convergence and mixing, the enhanced method is superior to (original) Structure MCMC and nearly as efficient as Order MCMC. Finally, Ellis and Wong (2008) propose the application of Order MCMC but followed by a correction step: for each sampled order, some number of consistent DAGs are sampled efficiently as described already in Friedman and Koller (2003) but attached with a term for correcting the bias due to the order-modular prior. However, as computing the correction term is computationally very demanding (#P-hard), approximations are employed.

The second concern in the original order MCMC scheme is that, while it greatly improves mixing compared to Structure MCMC, it may still get trapped at regions of locally high posterior probability. This deficiency of Order MCMC can be attributed to its particularly simple proposal distribution that induces a multimodal posterior landscape in the space of orders. Naturally, more sophisticated MCMC techniques can be successfully implemented upon the basic order MCMC scheme, as demonstrated by Ellis and Wong (2008). Such techniques would, of course, enhance the performance of Structure MCMC as well; see, e.g., Corander et al. (2008). Nevertheless, even if order based approaches are expected to be more efficient than DAG based approaches, it seems clear that multimodality of the posterior landscapes cannot be completely avoided. In this light, it is an intriguing question whether there are other sample spaces that yield still substantially improved convergence and mixing compared to the space of DAGs or linear orders.

Here, we answer this question in the affirmative. Specifically, we introduce *Partial Order MCMC* that samples partial orders on the nodes instead of linear orders. The improved efficiency of Partial Order MCMC stems from a conceptually rather simple observation: the computations per sampled partial order can be carried out essentially as fast as per sampled linear order, as long as the partial order is sufficiently "thin"; thus Partial Order MCMC benefits from a still much smaller sample space and smoother posterior landscape for "free" in terms of computational complexity. Technically, the fast computations per partial order rely on recently developed, somewhat involved dynamic programming techniques (Koivisto and Sood, 2004; Koivisto, 2006; Parviainen and Koivisto, 2010); we will review the key assumptions and results in Section 2. Regarding the sampler, in the present work we restrict ourselves to a handy, yet flexible class of partial orders, called *parallel bucket orders*, and to very simple Metropolis–Hastings proposals as analogous to Friedman and Koller's (2003) original Order MCMC. Furthermore, we focus on the posterior probabilities of individual arcs. These choices allow us to demonstrate the efficiency of Partial Order MCMC by comparing it to Order MCMC in as plain terms as possible.

How should we choose the actual family of partial orders, from which samples will be drawn? We may view this as a question of trading runtime against the size of the sample space. In Section 4 we present two observations. On one hand, our calculations suggest that, in general, the sample space should perhaps consist of singleton bucket orders rather than parallel compositions of several bucket orders. On the other hand, we show that one should operate with fairly large bucket sizes rather than buckets of size one, i.e., linear orders; it is shown how the reasonable values of the bucket size parameter depend on the number of nodes and the maximum indegree parameter.

Aside these analytical results, we study the performance of our approach also empirically, in Section 5. We show cases where Partial Order MCMC has substantial advantages over Order MCMC. The Markov chain over partial orders is observed to mix and converge much faster and more reliably than the chain over linear orders. Implications of these differences to structure discovery are illustrated by resulting deviations of the estimated arc posterior probabilities either as compared to exactly computed values or between multiple independent runs of MCMC.

## 2 PRELIMINARIES

### 2.1 BAYESIAN NETWORKS

A *Bayesian network* (BN) is a structured representation of a joint distribution of a vector of random variables $D = (D_1, \dots D_n)$. The structure is specified by a directed acyclic graph (DAG) $(N, A)$, where node $v \in N = \{1, ..., n\}$ corresponds to the random variable $D_v$, and the arc set $A \subseteq N \times N$ specifies the *parent set* $A_v = \{u : uv \in A\}$ of each node $v$; it will be notationally convenient to identify the DAG with its arc set $A$. For each node $v$ and its parent set $A_v$, the BN specifies a local conditional distribution $p(D_v|D_{A_v}, A)$. The joint distribution of $D$ is then composed as the product

$$p(D|A) = \prod_{v \in N} p(D_v|D_{A_v}, A) .$$

Our notation anticipates the Bayesian approach to learn $A$ from observed values of $D$, called *data*, by treating $A$ also as a random variable; we will consider this in detail in Section 2.3.

## 2.2 PARTIAL ORDERS

We will need the following definitions and notation.

A relation $P \subseteq N \times N$ is called a *partial order* on *base set* $N$ if $P$ is reflexive ($\forall u \in N : uu \in P$), antisymmetric (if $uv \in P$ and $vu \in P$ then $u = v$) and transitive (if $uv \in P$ and $vw \in P$ then $uw \in P$) in $N$. If $uv \in P$ we may say that $u$ *precedes* $v$ in $P$. A partial order $L$ on $N$ is a *linear order* or *total order* if, in addition, totality holds in $L$, that is, for any two elements $u$ and $v$, either $uv \in L$ or $vu \in L$. A linear order $L$ is a *linear extension* of $P$ if $P \subseteq L$.

A DAG $A$ is said to be *compatible* with a partial order $P$ if there exists a partial order $Q$ such that $A \subseteq Q$ and $P \subseteq Q$.

A family of partial orders $\mathcal{P}$ is an *exact cover* of the linear orders on $N$ if every linear order on $N$ is a linear extension of exactly one partial order $P \in \mathcal{P}$.

We say that a set $I \subseteq N$ is an *ideal* of a partial order $P$ on $N$ if from $v \in I$ and $uv \in P$ follows that $u \in I$. We denote the set of all ideals of $P$ by $\mathcal{I}(P)$.

In this paper we concentrate on a special class of partial orders known as *bucket orders*. Formally, let $B_1, B_2, \ldots, B_\ell$ be a partition of $N$ into $\ell$ pairwise disjoint subsets. A bucket order denoted by $B_1 B_2 \cdots B_\ell$ is a partial order $B$ such that $uv \in B$ if and only if $u \in B_i$ and $v \in B_j$ with $i < j$ or $u = v$. Intuitively, the order of elements in different buckets is determined by the order of buckets in question, but the elements inside a bucket are incomparable. Bucket order $B_1 B_2 \cdots B_\ell$ is said to be of length $\ell$ and of type $|B_1| * |B_2| * \ldots * |B_\ell|$. Partial order $P$ is a parallel composition of bucket order, or shortly *parallel bucket order*, if $P$ can be partitioned into $r$ bucket orders $B^1, B^2, \ldots B^r$ on disjoint basesets.

Bucket orders $B$ and $B'$ are *reorderings* of each other if they have the same baseset and are of same type. Further, parallel bucket orders $P$ and $\tilde{P}$ are reorderings of each other if their bucket orders can be labeled $B^1, B^2, \ldots, B^r$ and $\tilde{B}^1, \tilde{B}^2, \ldots, \tilde{B}^r$ such that $B^i$ and $\tilde{B}^i$ are reorderings of each other for all $i$. It is known that the reorderings of a parallel bucket order $P$, denoted by $\mathcal{R}(P)$, form an exact cover of the linear orders on their baseset (Koivisto and Parviainen, 2010).

It is well-known (see, e.g., Steiner, 1990) that a bucket order of type $b_1 * b_2 * \ldots * b_\ell$ has $\sum_{i=1}^{\ell} 2^{b_i} - \ell + 1$ ideals and $(b_1 + b_2 + \ldots b_\ell)!/(b_1! b_2! \cdots b_\ell!)$ reorderings. Furthermore, if $P$ is a parallel composition of bucket orders $B^1, B^2, \ldots, B^r$, it has $|\mathcal{I}(B^1)||\mathcal{I}(B^2)| \cdots |\mathcal{I}(B^r)|$ ideals and $t_1 t_2 \cdots t_r$ reorderings, where $t_i$ is the number of reorderings of $B^i$.

## 2.3 FROM PRIOR TO POSTERIOR

For computational convenience, we assume that the prior for the network structure $p(A)$ is *order-modular*, that is, in addition to the structure $A$, on the background there exists a hidden linear order $L$ on the nodes and the joint probability of $A$ and $L$ factorizes as

$$p(A, L) = \prod_{v \in N} \rho_v(L_v) q_v(A_v),$$

where $\rho_v$ and $q_v$ are some non-negative functions and $p(A, L) = 0$ if $A$ is not compatible with $L$. The prior for $A$ is obtained by marginalizing over $L$, that is, $p(A) = \sum_L p(A, L)$. Similarly, the prior for $L$ is $p(L) = \sum_A p(A, L)$. For our purposes, it is essential to define a prior for every partial order $P \in \mathcal{P}$, where $\mathcal{P}$ is an exact cover of the linear orders on $N$. We get the prior by marginalizing $p(L)$ over the linear extensions of $P$, that is, $p(P) = \sum_{L \supseteq P} p(L)$.

Our goal is to compute posterior probabilities of modular structural *features* such as arcs. To this end, it is convenient to define an indicator function $f(A)$ which returns 1 if $A$ has the feature of interest and 0 otherwise. A feature $f(A)$ is *modular* if it can be expressed as a product of local indicators, that is, $f(A) = \prod_{v \in N} f_v(A_v)$. For example, an indicator for an arc $uw$ can be obtained by setting $f_w(A_w) = 1$ if $u \in A_w$, $f_w(A_w) = 0$ otherwise, and $f_v(A_v) = 1$ for $v \neq w$. For notational convenience, we denote the event $f(A) = 1$ by $f$.

Now, the joint probability $p(D, f, P)$, which will be a key term in the computation of the posterior probability of the feature $f$, can be obtained the following manner. First, by combining the order-modular prior and the likelihood of the DAG, we get

$$p(A, D, L) = p(A, L) p(D|A).$$

Furthermore, for $P \in \mathcal{P}$, we get

$$p(A, D, P) = \sum_{L \supseteq P} p(A, L) p(D|A).$$

Finally, we notice that the feature $f$ is a function of $A$ and thus

$$p(D, f, P) = \sum_{L \supseteq P} \sum_{A \subseteq L} f(A) p(A, L) p(D|A).$$

The order modularity of the prior and the modularity of the feature and likelihood allows us to use the algorithms of Parviainen and Koivisto (2010) and thus conclude that if each node is allowed to have at most $k$ parents then $p(D, f, P)$ can be computed in time $O(n^{k+1} + n^2 |\mathcal{I}(P)|)$. In fact the same bound holds for the computation of the posterior probability of all arcs.

# 3 PARTIAL ORDER MCMC

We can express the posterior probability of the feature $f$ as the expectation of $p(f|D, P)$ over the posterior of the partial orders:

$$p(f|D) = \sum_{P \in \mathcal{P}} p(P|D)p(f|D, P) \,.$$

Because the exact covers $\mathcal{P}$ we consider are too large for exact computation, we resort to importance sampling of partial orders: we draw partial orders $P_1, P_2, \ldots, P_T$ from the posterior $p(P|D)$ and estimate

$$p(f|D) \approx \frac{1}{T} \sum_{i=1}^{T} p(f|D, P_i) \,.$$

Since $p(f|P, D) = p(D, f, P)/p(D, P)$, we only need to compute $p(D, f, P)$ and $p(D, P)$ up to a (common) constant factor.

Because direct sampling of the posterior $p(P|D)$ seems difficult, we employ the popular MCMC method and sample the partial orders along a Markov chain whose stationary distribution is $p(P|D)$. We next construct a valid sampler for any state space, i.e., family of partial orders, that consists of the reorderings of a parallel bucket order. For concreteness, we let $P^*$ be a parallel composition of $r$ bucket orders $B^1, B^2, \ldots, B^r$, each of type $b * b * \ldots * b$ and, thus, length $n/(br)$. Now, the state space is $\mathcal{R}(P^*)$.

We use the Metropolis–Hastings algorithm. We start from a random initial state and then at each *iterations* either move to a new state or stay at the same state. At a state $P$ a new state $\tilde{P}$ is drawn from a suitable proposal distribution $q(\tilde{P}|P)$. The move from $P$ to $\tilde{P}$ is accepted with probability

$$\min \left\{ 1, \frac{p(\tilde{P}, D)q(P|\tilde{P})}{p(P, D)q(\tilde{P}|P)} \right\}$$

and rejected otherwise.

While various proposal distributions are possible, we here focus on a particularly simple choice that considers all possible node pair flips between two distinct buckets within a bucket order. Formally, we select an index $k \in \{1, ..., r\}$ and a node pair $(u, v) \in B_i^k \times B_j^k$, with $i < j$, uniformly at random, and construct $\tilde{P}$ from $P$ by replacing $B^k$ by $\tilde{B}^k$, where

$$B^k = B_1^k \ldots B_i^k \ldots B_j^k \ldots B_l^k \quad \text{and}$$
$$\tilde{B}^k = \tilde{B}_1^k \ldots \tilde{B}_i^k \setminus \{u\} \cup \{v\} \ldots \tilde{B}_j^k \setminus \{v\} \cup \{u\} \ldots \tilde{B}_l^k .$$

To see that the algorithm works properly, that is, the Markov chain is ergodic and its stationary distribution is $p(P|D)$, it is sufficient to notice that all the states are accessible from each other, the chain is aperiodic, and $p(P, D)$ is proportional to $p(P|D)$.

The time requirement of one iteration is determined by the complexity of computing $p(P, D)$ up to a constant factor. As already mentioned, this time requirement is $O(n^{k+1} + n^2|\mathcal{I}(P^*)|)$, where $k$ is the maximum indegree. If we restrict ourselves to the described node flip proposals, this bound can be reduced using the trick of Friedman and Koller (2003) to $O(n^k + n^2|\mathcal{I}(P^*)|)$; we omit the details due to lack of space.

# 4 ANALYTICAL RESULTS

Our description of Partial Order MCMC in the previous section left open the precise choice of the family of parallel bucket orders. Next, we derive guidelines for choosing "optimal" values for the associated parameters: the number of parallel bucket orders, $r$, and the bucket size in each bucket order, $b$; for simplicity we do not consider settings where bucket sizes differ within a bucket order or between bucket orders. In what follows, $P$ will be a composition of $r$ parallel bucket orders of type $b * b * \cdots * b$ and length $\ell$. Recall that $P$ specifies $\mathcal{R}(P)$, the reorderings of $P$, which is the sample space of Partial Order MCMC for problems on $n = r\ell b$ nodes. Note that when $r = 1$ and $b = 1$, the family $\mathcal{R}(P)$ is the family of all linear orders on the $n$ nodes, that is, the case of Order MCMC.

The choice of the parameters of $P$ is guided by two contradictory goals: both the number of ideals of $P$ and the size of $\mathcal{R}(P)$ should be as small as possible. The former determines the runtime of a single MCMC iteration, and is given by $(\ell 2^b - \ell + 1)^r$. The latter determines the size of the sample space, and is given by $\big((\ell b)!/(b!)^\ell\big)^r$. In the following paragraphs we address two questions: *Is it always most reasonable to have just one (parallel) bucket order, that is, $r = 1$? If $r = 1$, how large can the bucket size $b$ be still guaranteeing that the runtime becomes only negligibly larger compared to the case of $b = 1$?*

For the first question, we give a rough calculation. With large enough $b$, we may estimate the size of the sample space by $(\ell b/e)^{\ell b r}/(b/e)^{b\ell r} = \ell^n$, by using Stirling's approximation, $t! \approx (t/e)^t$. Thus, this dominating factor in the sample space matches for different values of $r$ when the length of the bucket orders are equal and the products $br$ are equal. Now, since the dominating term in the runtime is $\ell^r 2^{br}$, we conclude that increasing $r$ increases the runtime, when the size of the sample space is (approximately) fixed.

**Observation 1** *Having more than one (parallel) bucket order is not likely to yield substantial advan-*

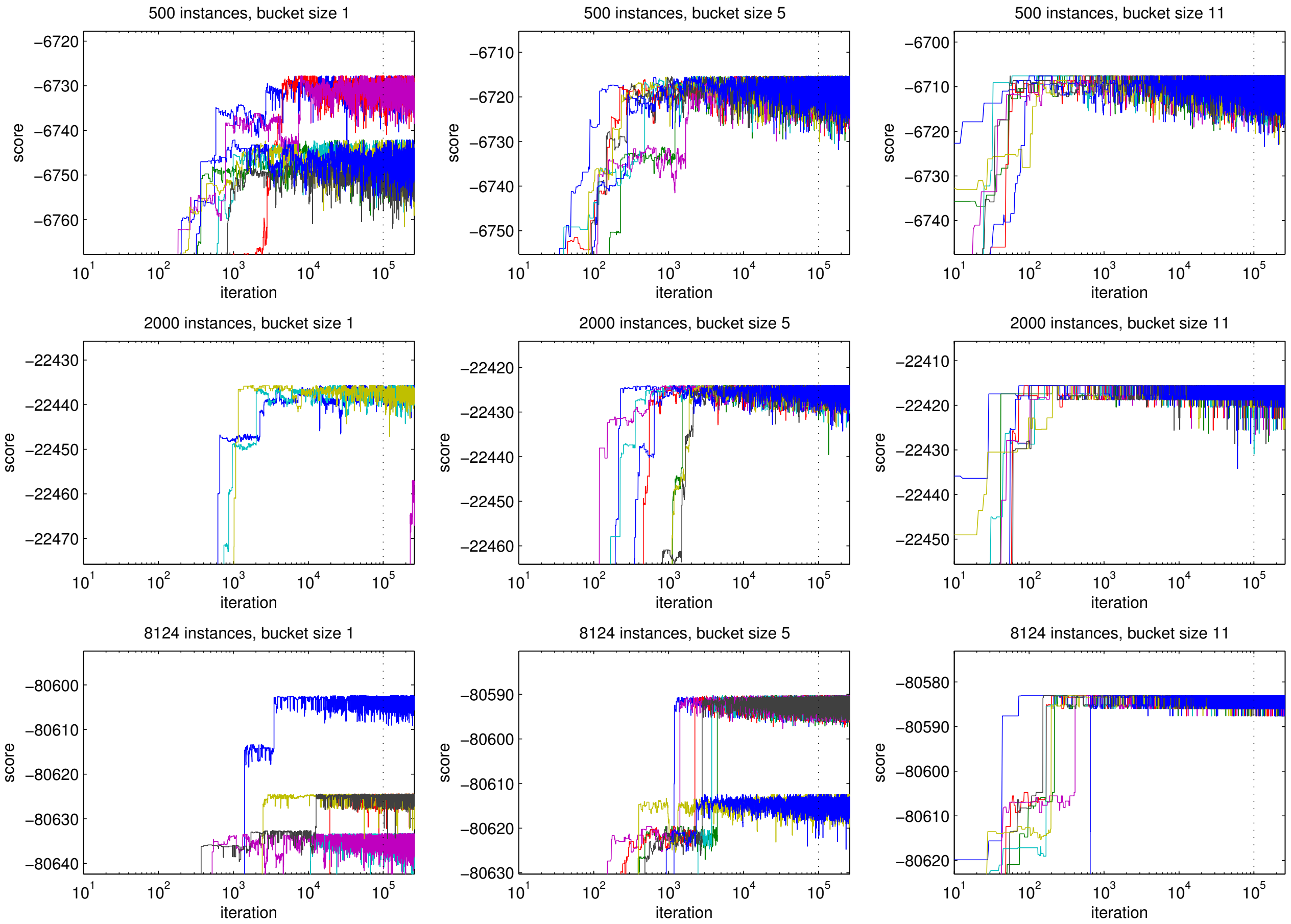

Figure 1: Convergence of MCMC for different number of instances and bucket sizes for Mushroom. Each panel shows the evolution of eight runs of MCMC with random starting states. (Some runs got stuck in very low probability states and are not visible at all.) Scores, that is, the logarithms of unnormalized state probabilities after each MCMC step are plotted on y-axis. Note, that since bucket orders with different bucket sizes cover different numbers of linear orders, the scores are not directly comparable between the panels. The dotted vertical lines indicate the point where the burn-in period ended and the actual sampling started.

*tages in the runtime, assuming the size of the sample space is fixed.*

For the second question, we refer to the time requirement $O(n^{k+1}+n^2 2^b n/b)$ per MCMC iteration (for general proposal distributions) and notice that the latter term has only a negligible influence on the bound as long as $2^b/b \leq n^{k-2}$.

**Observation 2** *Having bucket sizes larger than one are likely to yield substantial advantages in the sample space size, assuming the runtime per MCMC iteration is fixed. A reasonable bucket size is* $b \approx (k-2)\log_2 n$.

Compared to linear orders, the size of the sample space reduces by an exponential factor: $b!^{n/b} \approx (b/e)^n$.

# 5 EMPIRICAL RESULTS

We have implemented the presented algorithm in C++. The implementation includes the exact computation of $p(D, f, P)$ for arc features and a Partial Order MCMC restricted to bucket orders with equal bucket sizes (the last bucket is allowed to be smaller in case of nondivisibility). The software is available at `http://www.cs.helsinki.fi/u/tzniinim/bns/`.

## 5.1 DATASETS AND PARAMETERS

In our experiments we used two datasets: Mushroom (Frank and Asuncion, 2010) contains 22 discrete attributes and 8124 instances. Alarm (Beinlich et al., 1989) is a Bayesian network of 37 nodes from which we sampled independently 10 000 instances. In addition, we also considered random samples of 500 and 2000

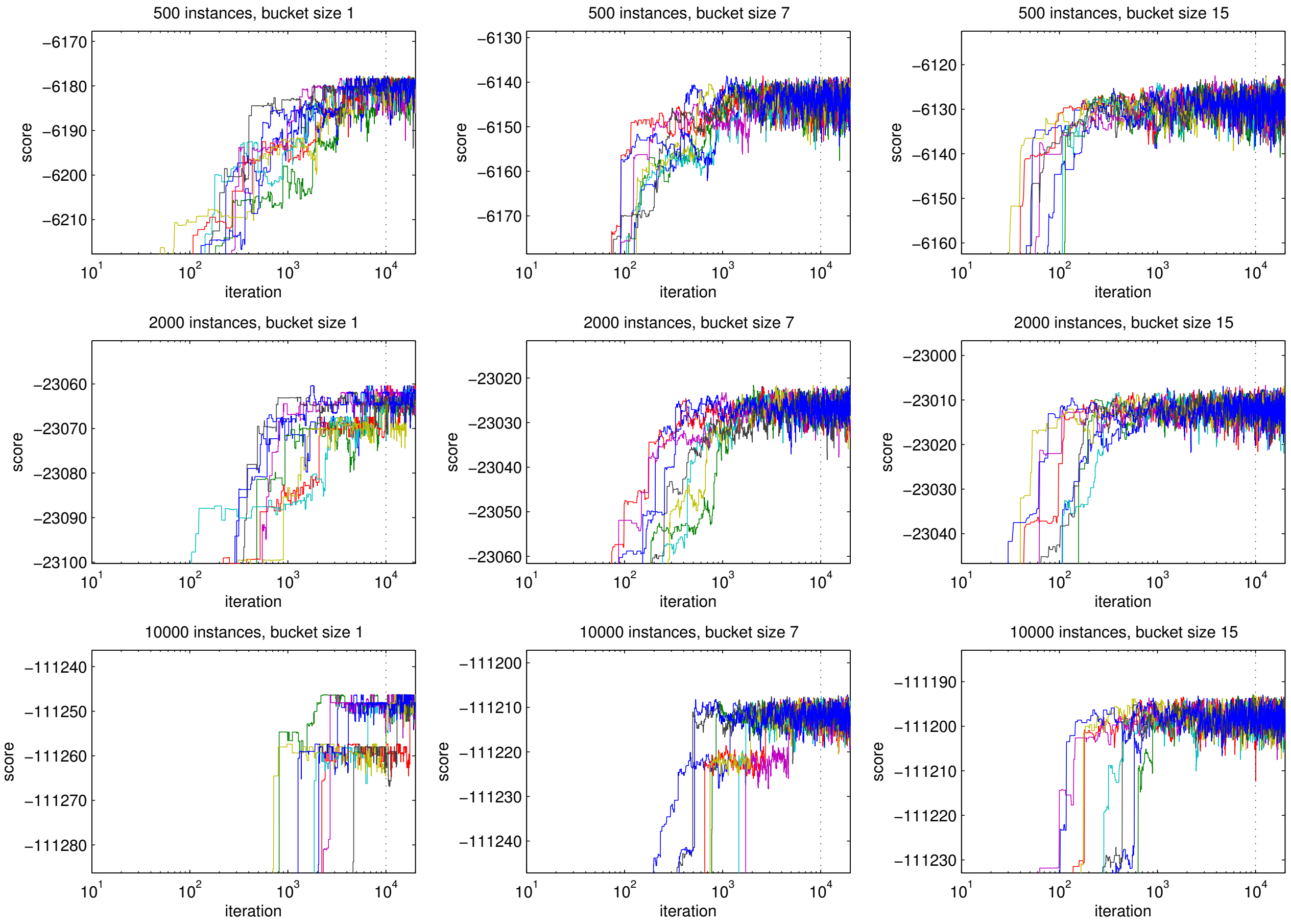

Figure 2: Mixing and convergence of MCMC for ALARM. For further explanation, see Figure 1.

instances from both data sets.

The conditional probabilities $p(D_v|D_{A_v}, A)$ were assigned the so-called K2 scores (Cooper and Herskovits, 1992). For the order-modular prior we set $\rho_v(L_v) = 1$ and $q_v(A_v) = 1/\binom{n-1}{|A_v|}$ for all $v$, $L_v$ and $A_v$.

For MUSHROOM we set the maximum indegree $k$ to 5 and let the bucket size vary from 1 to 11. We first ran the sampler 100 000 steps for "burn-in", and then took 400 samples at intervals of 400 steps, thus, 260 000 steps in total. All arc probabilities were estimated based on the collected 400 samples. Due to the relatively low number of nodes, we were also able to compute the exact arc probabilities for comparison.

For ALARM we fixed the maximum indegree to 4 (which matched the data-generating network) and let the bucket size vary from 1 to 15. We ran the sampler 10 000 steps for burn-in, and then took 100 samples at intervals of 100 steps, thus, 20 000 steps in total.

For both datasets and each parameter combination (number of instances, bucket size), we conducted eight independent MCMC runs from random starting states.

## 5.2 MIXING AND CONVERGENCE

For reliable probability estimates, it is important that the Markov chain mixes well and the states with high probability are visited. If the algorithm gets trapped in a local region with low probability the resulting probability estimates can be very inaccurate. We studied how the mixing rate is affected by bucket size by inspecting the evolution of the Markov chain state probability $p(P|D)$ in the eight independent runs.

For MUSHROOM the results for three different bucket sizes are shown in Figure 1. We observe that the chains with larger buckets mix significantly faster than the chains with smaller buckets: With bucket size one—which directly corresponds to Order MCMC—all runs do not converge to the same probability levels; some runs are trapped in states of much lower probabilities than some other runs. However, increasing the bucket size enables all runs not only to converge to the same probability levels but also to do it in fewer steps. This phenomenon is observed for each of the three data sizes, albeit the convergence is notably faster for smaller numbers of instances.

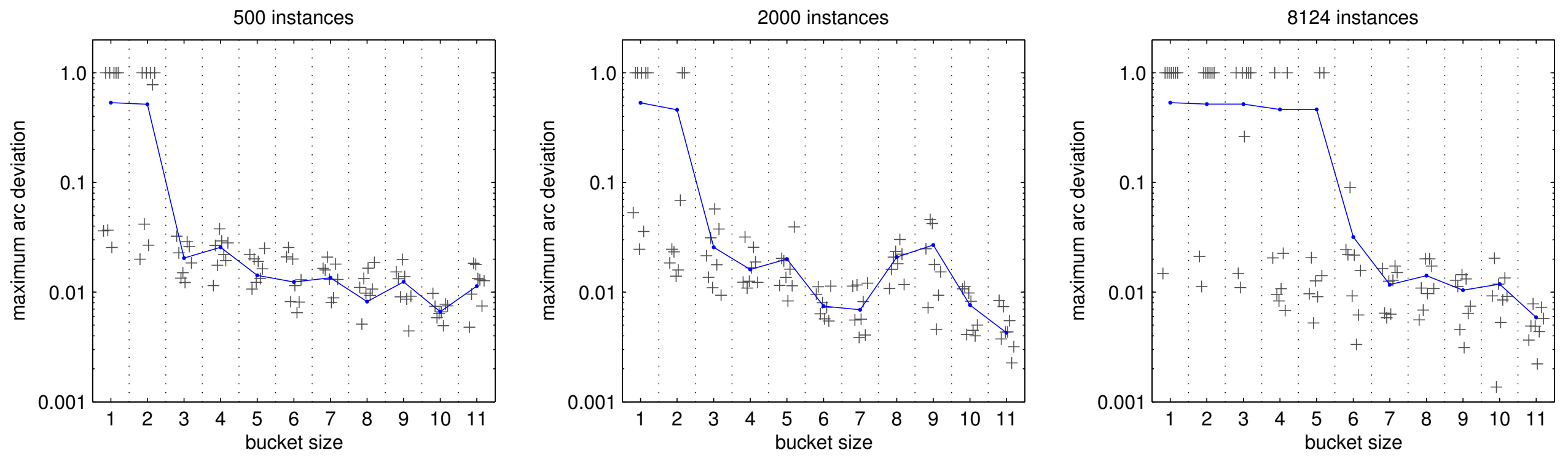

(a) MUSHROOM

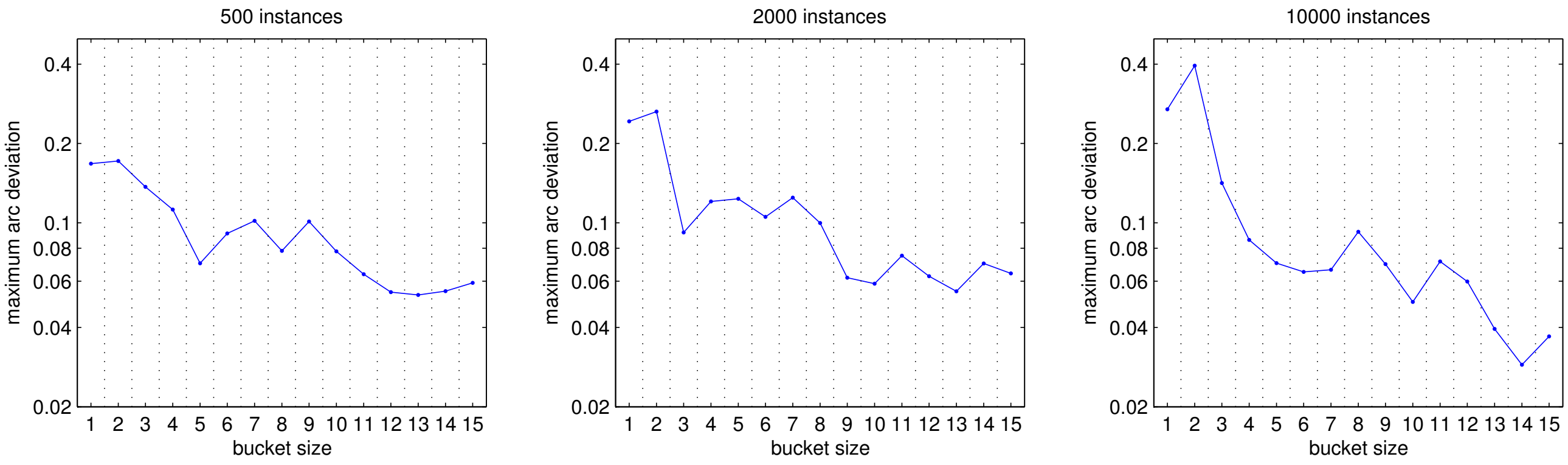

(b) ALARM

Figure 3: Maximum arc deviations for MUSHROOM and ALARM. The solid curve indicates the maximum arc deviation as a function of bucket size. In addition, for MUSHROOM, for each bucket size and each of eight runs the largest absolute error is plotted as +. Values are slightly perturbed horizontally to ease visualization.

For ALARM the respective evolution of state probabilities is shown in Figure 2. Again we observe the same tendencies as we did for MUSHROOM. Interestingly, however, ALARM appears to be an easier dataset: even though the the number of MCMC iterations is an order of magnitude smaller than for MUSHROOM, all runs seem to eventually convergence, also with small bucket sizes. Yet, the runs with bucket size 1 still need about 10–100 times more iterations than the runs with bucket size 15.

For both MUSHROOM and ALARM the acceptance ratio of Metropolis-Hastings sampler—that is, the proportion of state transition proposals which were accepted—was between 0.05 and 0.4. Increasing of the maximum indegree, and to a lesser extent increasing of the bucket size, had a clear but not dramatic negative effect on the acceptance ratio. (Data not shown.)

### 5.3 VARIANCE OF ESTIMATES

We measured the accuracy of arc probability estimates as follows. For each arc we calculated the standard deviation among the eight estimates, one from each of the eight MCMC runs. Then, the largest of these deviations, which we call the *maximum arc deviation*, was used as the measure. In addition, for MUSHROOM we were able to measure the accuracy of a single MCMC run by the *largest absolute error*: $\max_a |\hat{p}_a - p_a|$, where $a$ ranges over different arcs, $p_a$ is the exact probability of $a$ and $\hat{p}_a$ is the respective MCMC estimate.

The results for MUSHROOM are shown in Figure 3a. Generally, the accuracy seems to improve when the bucket size increases. This is mainly due to some runs with small bucket sizes that estimate the probability of some arcs completely opposite to the exact probability. However, even if we ignore these cases, we see a clear tendency for more accurate estimates as the bucket size increases.

Maximum arc deviations for ALARM are shown in Figure 3b. Again the accuracy seems to improve as the bucket size grows, though the trend is more subtle than for MUSHROOM. Yet, even with the easiest case of 500 instances, the maximum arc deviation is about 0.2 for bucket size 1, but just about 0.06 for bucket sizes 12 or larger.

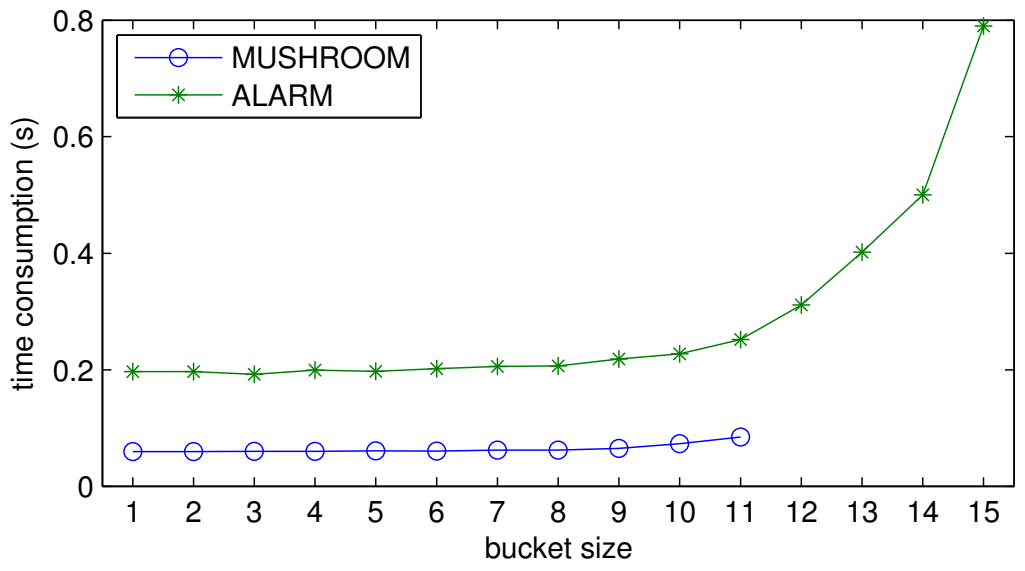

Figure 4: Time consumption of one MCMC iteration as a function of bucket size for MUSHROOM and ALARM.

### 5.4 TIME CONSUMPTION

The time consumption of one MCMC iteration is shown in Figure 4. The observations are in good agreement with the asymptotic time requirement $O(n^{k+1} + n^3 2^b/b)$: for small bucket sizes the number of possible parent sets dominates, but for larger bucket sizes, the time consumption starts to grow exponentially. For both MUSHROOM and ALARM, the fixed maximum indegrees, 5 and 4, respectively, allow having the bucket size up to about 10, still with essentially no penalty in the running time. This agrees well with the rough guideline of Observation 2 that suggests bucket sizes about $(5-2)\log_2 22 \approx 13.4$ and $(4-2)\log_2 37 \approx 10.4$ for MUSHROOM and ALARM, respectively.

## 6 CONCLUDING REMARKS

We presented a new Partial Order MCMC algorithm for estimating probabilities of structural features in Bayesian Networks. The algorithm was implemented and compared to Order MCMC with favourable results. This basic version of Partial Order MCMC readily enables upgrading just like Order MCMC as described by Ellis and Wong (2008): (a) for correcting the "prior bias" and estimating the posterior of more complex structural features by sampling DAGs compatible with an order, and (b) for further enhancing mixing and convergence by more sophisticated MCMC techniques.

### Acknowledgements

This research was supported in part by the Academy of Finland, Grant 125637 (M.K.).